\pdfoutput=1

\documentclass[11pt]{article}
\usepackage{amsmath}
\usepackage{multirow}
\usepackage[inline]{enumitem}
\usepackage{enumerate}  
\usepackage{amsfonts}
\usepackage[dvipsnames, svgnames, x11names]{xcolor} 
\usepackage{booktabs}
\usepackage{ACL2023}
\usepackage{xspace}
\usepackage{times}
\usepackage{latexsym}
\usepackage{graphicx} 
\usepackage{amssymb}   
\usepackage{colortbl}
\usepackage[T1]{fontenc}
\usepackage{verbatim}
\usepackage[utf8]{inputenc}
\usepackage{makecell}
\usepackage{microtype}
\usepackage{pifont}

\usepackage{inconsolata}
\usepackage{xcolor,soul}
\usepackage[toc,page]{appendix}
\title{RADE: Reference-Assisted Dialogue Evaluation for Open-Domain Dialogue}

\definecolor{lemon}{HTML}{FDFFCC}

\newcommand{\hllemon}{}

\author{
Zhengliang Shi$^1$, Weiwei Sun$^1$, Shuo Zhang$^2$, Zhen Zhang$^1$, \\
{\bf Pengjie Ren$^1$, Zhaochun Ren\textsuperscript{$1$}\thanks{$^*$ Corresponding author.} } \\
$^1$Shandong University, Qingdao, China  \\  
$^2$Bloomberg, London, United Kingdom \\
\texttt{
shizhl@mail.sdu.edu.cn~~\{sunnweiwei, zhen.zhang.sdu\}@gmail.com}
\\
\texttt{
zhaochun.ren@sdu.edu.cn~~szhang611@bloomberg.net~~jay.ren@outlook.com}
}

\newcommand{\code}[1]{{\ttfamily#1}}

\begin{document}
\maketitle


\begin{abstract}

Evaluating open-domain dialogue systems is challenging for reasons such as the one-to-many problem, i.e., many appropriate responses other than just the golden response.   
As of now, automatic evaluation methods need better consistency with humans, while reliable human evaluation can be time- and cost-intensive. 
To this end, we propose the \textbf{R}eference-\textbf{A}ssisted  \textbf{D}ialogue \textbf{E}valuation (RADE) approach under the multi-task learning framework, which leverages the pre-created utterance as reference other than the gold response to relief the one-to-many problem. 
Specifically, RADE explicitly compares reference and the candidate response to predict their overall scores.
Moreover, an auxiliary response generation task enhances prediction via a shared encoder.
To support RADE, we extend three datasets with additional rated responses other than just a golden response by human annotation.
Experiments on our three datasets and two existing benchmarks demonstrate the effectiveness of our method, where Pearson, Spearman, and Kendall correlations with human evaluation outperform state-of-the-art baselines.

\end{abstract}

\section{Introduction}\label{sec:intro}

\begin{figure}[t]
        \centering
	\includegraphics[width=1\linewidth]{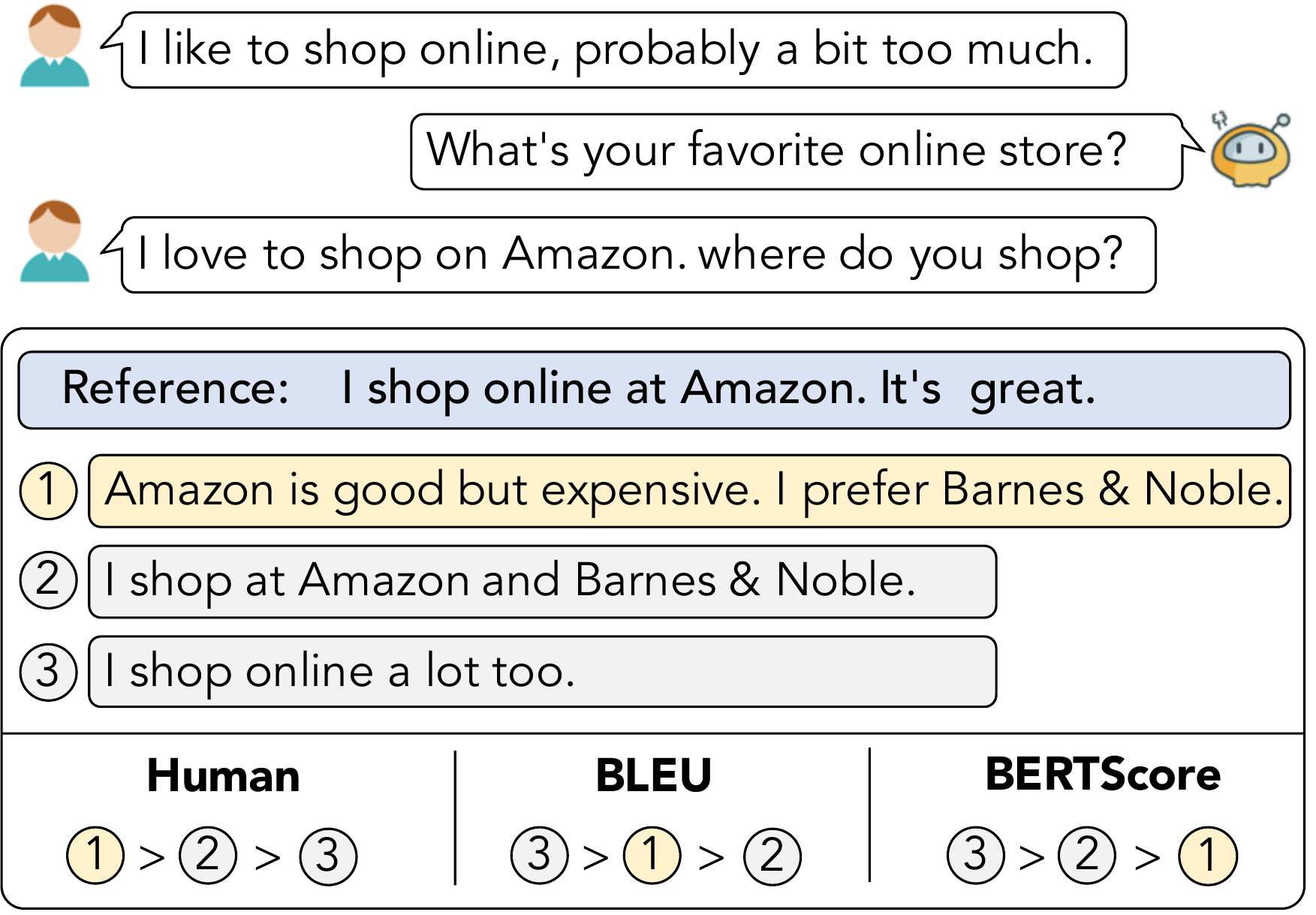}
        \caption{An example to explain the one-to-many nature of open-domain dialogues. }
 \label{intro:case}
\end{figure}

Open-domain dialogue system, which focuses on non-goal-oriented chitchat, may converse on a broad range of arbitrary topics. Recent years have witnessed rapid advances in natural language generation~\citep{zhang2019dialogpt,blenderbot,LLMSurvey}, boosting the development of open-domain dialogue systems. 
Conversations with such systems resemble human-human interactions as various responses might fit the context, given that users often do not have a specific goal beyond enjoying the conversation. Evaluating these conversations is thus challenging because of the so-called one-to-many problem~\citep{ZhangmingChan2021EnhancingTO,ji2022achieving}; see Figure~\ref{intro:case} where three candidate responses with different semantics fit the context while there is only one golden response.

The most common practice of dialogue evaluation is done with reference-based metrics, which compare the generated response with a pre-created response, commonly referred to as the golden standard~\citep{ji2022achieving}.
The reference-based metrics calculate the similarity between the generated and gold responses at either lexical level (e.g., ROUGE ~\citep{lin-2004-rouge}, BLEU~\citep{papineni2002bleu}) or semantic level (e.g., BERTScore~\citep{TianyiZhang2019BERTScoreET}, ADEM~\citep{RyanLowe2017TowardsAA}). 
However, these metrics ignore the one-to-many nature of open-domain dialogues.
As illustrated at the bottom of Figure~\ref{intro:case}, the generated response ``\emph{Amazon is good but expensive ...}'' expresses the opposite semantics to the golden response ``\emph{I shop online...}'' and is therefore considered a non-good response by the reference-based metrics.
Therefore, these metrics may need a higher consistency with humans.
Recently, \textit{multi-reference methods} and \textit{reference-free methods} are proposed to address the drawback of reference-based metrics.
The former explicitly annotates multiple references for dialogue~\citep{eric2021multi}, whereas the latter discards the golden response in the evaluation and achieves high correlations with human judgments~\citep{mehri2020usr,LishanHuang2020GRADEAG}.
However, drawback still exists in these two classes of methods.
Multi-reference methods are costly and hard to generalize to different datasets, while reference-free methods are often unstable and vulnerable to data-induced biases\footnote{The data-induced biases included two aspects: (1) Noise collected in data/annotations, (2)  The reference-free models tend to favor the underlying models' outputs and those from similar models or trained with similar datasets.~\citep{Khalid2022ExplainingDE,Deutsch2022OnTL}}.

To overcome the weakness of existing evaluation methods and further resolve the one-to-many problem, we propose a new technique, namely \textbf{R}eference-\textbf{A}ssisted \textbf{D}ialogue \textbf{E}valuation (RADE).
RADE considers the pre-created response as a reference instead of the golden standard.

To support RADE, we design a new human annotation task to extend existing datasets, which includes metric decompose and pairwise annotation, where a pre-scored golden response is paired with generated responses for rating following a unified rating score. The final scores are arrived at by aggregating ratings with a weighted sum from different sub-metrics. The human annotation collects labels for three high-quality datasets with 10,112 dialogues, which correspond to three downstream open-domain dialogue system tasks, i.e., chitchat, empathetic dialogue, and personal chat. These multi-domain datasets make RADE more robust when generalizing to cross-domain evaluation scenarios while having a better task-specific performance.

We propose a RADE model under the multi-task learning framework for automatic evaluation based on the newly collected datasets. 
Specifically, RADE first explicitly encodes the relation between dialogue context and generated response with reference assistance. Then RADE discriminates whether the reference or response fits the context better and predicts the scores for each utterance. To relieve the one-to-many problem, we augment RADE with a joint response generation task where RADE learns to generate the reference responses to better perceive the range of candidate responses.

Extensive experiments on our three benchmarks demonstrate that RADE achieves the best correlations with human judgment.
We also examine two existing USR benchmark~\citep{mehri2020usr} where RADE outperforms the state-of-the-art methods, e.g., pushing the Pearson correlation coefficient to 48\% (6.8\% absolute improvement) and Spearman correlation coefficient to 46.6\% (4.3\% absolute improvement).
Experiments also verify the generalizability of our proposed method.

Our contributions can be summarized as follows:
(1) We propose the reference-assisted evaluation method, i.e., RADE, for open-domain dialogue evaluation; 
(2) We design a new human annotation task 
and collect three new dialogue evaluation datasets;
(3) Experiments on our benchmarks and two existing benchmarks verify the effectiveness and robustness of the proposed methods; 
(4) We release three new benchmarks and the pre-trained evaluation model to facilitate future research on dialogue evaluation.

\section{Related work}\label{sec:relatedwork}

\subsection{Reference-based dialogue evaluation}

Previous reference-based methods compare the generated response with the pre-created response at the lexical or semantic level.
Lexical-level metrics, e.g., ROUGE~\citep{lin-2004-rouge}, BLEU~\citep{papineni2002bleu} and METEOR ~\citep{banerjee2005meteor}, count the n-gram overlap between the candidate response and the reference response.
These methods usually correlate poorly with human evaluation results due to the lexical mismatch problem~\citep{ChiaWeiLiu2016HowNT}.
Semantic-level metrics evaluate address lexical mismatch problem by calculating similarity with high-dimension embeddings.
For example, \citet{nlg} measures the embedding distance between golden and generated response.
\citet{SarikGhazarian2019BetterAE} and \citet{TianyiZhang2019BERTScoreET} enhance the text representation using the large pre-train model, which has shown exemplary performance in capturing semantic similarity.
However, they suffer from the one-to-many problem when evaluating open-domain dialogues since responses with various semantics may fit the dialogue context.

Recent works tend to relieve this drawback by
annotating multiple references for dialogue, commonly referred to as multi-reference methods~\citep{li2017dailydialog, deb}, which are costly and hard to generalize to agnostic scenarios.
The proposed RADE aims to consider the pre-created response as a candidate instead of the golden standard to address the one-to-many problem of dialogue evaluation.

\subsection{Reference-free dialogue evaluation}
The reference-free methods are gaining more attention as they correlate more with human judgment only with the dialogue context and response.
For example, MAUDE predicts the score of dialogue using pre-trained language models,
GRADE~\citep{LishanHuang2020GRADEAG} evaluates the coherence of dialogues with the augmentation of the commonsense graph,
EMS~\citep{ZhangmingChan2021EnhancingTO} enhances the dialogue evaluation by capturing the representation of the context and response in latent space.
Some methods further decompose the evaluation of responses into multiple perspectives~\citep{ShikibMehri2020UnsupervisedEO,mehri2020usr,usl}, such as relevance, fluency, and engagingness, then aggregate the overall score from different sub-metrics with a weighted average.
However, some recent studies~\citep{Khalid2022ExplainingDE,Deutsch2022OnTL} reveal that the reference-free methods are vulnerable to data-induced biases and inherently biased toward models which are more similar to their own.
In contrast, this paper proposes a reference-assisted approach, which enhances the robustness of the model using reference responses as a benchmark.

\section{Task Formulation }\label{sec:task}

In this work, we propose two tasks: (1) extending the existing datasets by human annotation, and (2) leveraging the rated references collected in (1) to enhance automatic evaluation.

\paragraph{Human annotation}
Human annotation aims to extend existing datasets with multiple rated responses to facilitate automatic evaluation. 
Given a dialogue context $c$, which is always paired with a golden response (denoted as reference) $r_h$, we employ the generation models, e.g., BlenderBot~\citep{blenderbot}, to generate one more response $r_a$.
We then assign a fixed overall score or derive from existing datasets to the reference as $s_h$.
The annotators are instructed to rate $r_a$ as $s_a$, following the same scale while taking the reference as a benchmark.
The annotators are also asked to revise the reference score $s_h$ if $s_h$ is inappropriate.

\paragraph{Automatic evaluation}
Given a dialogue context $c$, the proposed RADE learns to evaluate the response $r_a$ with the assistance of reference $r_h$ under the multi-task learning framework.
The first task explicitly models the relation between reference and response and discriminates which fits the context better.
The scores of reference and response are predicted simultaneously.
And the second task enhances the score prediction task by implicitly estimating the distribution of candidate responses.

\section{Human Annotation}
\label{sec:ref-human}

Our human annotation task aims to rate the candidate responses following a pre-scored reference as a benchmark.
Since there are multiple perspectives to assess the response, we simplify by sorting the possible aspects into two categories: the general view and the task-specific view.
As listed in Table~\ref{metrics}, the former contains relevance, engagingness, and fluency, which are suitable for all dialogue agents.
And task-specific criteria consist of understandability, emotional awareness, and personality awareness, which correspond to chitchat dialogue, emotional dialogue, and persona dialogue.
We annotate rates on each metric and calculate the overall rating score by weighting these sub-metrics.
Specifically, the weights are obtained based on the preference of users (see section~\ref{user_study} for more details).

\begin{table}[!t]\small

\begin{tabular}{@{}p{7.7cm} ccc @{}}
\toprule
\textbf{Relevance $^\dagger$:}\\
\textit{Whether the response matches dialogue context semantically. } \\
\midrule
\textbf{Engagingness$^\dagger$:}\\
\textit{Whether the response is engaging or interesting rather than rigid template.} \\
\midrule

\textbf{Fluency$^\dagger$:}\\
\textit{Whether the response is fluent and natural throughout the conversation.}  \\
\midrule

\textbf{Understandability$^\ddagger$:}\\
\textit{Is there any external knowledge contained in the response.}  \\
\midrule

\textbf{Emotional-awareness$^\ddagger$:}\\
\textit{Whether the agent capture the emotion  of user and support empathic support.} \\
\midrule

\textbf{Personality-awareness$^\ddagger$:}\\
\textit{Whether the response conforms to given personality.}\\ 

\bottomrule
\end{tabular}
\caption{\textbf{Criteria in human annotation.}
Metrics with $^\dagger$ are general metrics for all dialogue tasks, while metrics $^\ddagger$ are metrics for specific dialogue tasks (e.g., understandability for chitchat, emotion-awareness for emotional dialogue and personal-awareness for personal chat).}
\label{metrics}
\end{table}

\subsection{Data preparation}

We consider three datasets to extend: 
\begin{itemize*}
    \item \emph{DSTC-ChitChat (ChitChat)}~\citep{hori2017end}, a chitchat dataset collected from Twitter,  each example derived from the conversation between a customer and an agent. 
    \item \emph{Empathetic Dialogues (EmpaDial)}~\citep{rashkin2019towards}, which consists of 25k dialogues grounded in emotional situations.
    \item \emph{PersonaChat}~\citep{zhang2018personalizing}, a real-world dataset consisting of 10k dialogues where each participant plays the part of an assigned persona.
\end{itemize*}

Then, we collect model-generated responses using the following seven well-performing dialogue models on these datasets: 
BlenderBot~\citep{blenderbot},
DialoGPT~\citep{zhang2019dialogpt},
KEMP~\citep{QintongLi2023KnowledgeBF},
MoEL~\citep{ZhaojiangLin2019MoELMO},
MIME~\citep{majumder2020mime},
EmpDG~\citep{QintongLi2019EmpDGMI},
PersonaGPT~\citep{tang2021persona}.

The train-dev-test of collected datasets are split as Chitchat (1490/300/300, 5/1/1), Empathetic Dialogue (3022/500/500, 6/1/1), and Persona Chat (3000/500/500, 6/1/1).
More details of these models are available in Appendix~\ref{models}.

\subsection{Human annotation detalis}

We hire 40 annotators for data annotation.
Following a five-scale standard, they are asked to label sub-metrics as listed in Table~\ref{metrics}. 
The five-scale allows the annotators to factor in their subjective interpretation of the extent of success or failure of a system's response to satisfy a user's request.
The dialogue context, rated reference response, and corresponding score are provided in each example.
At least three annotators are required for each example.
We annotated about 10k dialogues for the three datasets, and the statistics of the collected datasets are listed in Table~\ref{statis}.
The ratings achieve reasonable inter-annotator agreements with Fleiss Kappa scores of 0.540, 0.554, and 0.533 on three datasets, respectively.
More details about the annotation guideline and details are provided in Appendix~\ref{ui}.
\begin{table}[!t]
\centering\small

\setlength\tabcolsep{8pt}
\begin{tabular}{@{} l ccc @{}}
\toprule
\textbf{Domain} & \textbf{ChitChat} & \textbf{EmpaDial} & \textbf{PersonaChat} \\ 
\midrule
\# Dialogues &  2,090      & 4,022 & 4,000    \\
Kappa  &  0.540      & 0.554 & 0.533 \\ 
\midrule 
\multicolumn{4}{@{}l}{\emph{Distribution of the score}}  \\
Rating 1 & \phantom{0}0.5\%  & \phantom{0}1.2\% & \phantom{0}3.7\% \\
Rating 2 &15.6\% & 12.5\% & 12.6\% \\
Rating 3 & 48.3\% & 42.0\% & 50.5\%\\
Rating 4 & 29.5\% & 32.0\% & 23.9\% \\
Rating 5 & \phantom{0}5.1\%  & 12.3\% &  \phantom{0}9.4\%\\
\bottomrule
\end{tabular}
\caption{
\textbf{The statistics of the collected datasets.}
For each example, the overall score of the response is mean of all sub-metrics.
 }
\label{statis}
\label{annotation}
\end{table}

\section{Reference-Assisted Automatic Evaluation}\label{sec:ref-auto}

We propose RADE, a \textbf{R}eference-\textbf{A}ssisted Automatic \textbf{D}ialogue \textbf{E}valuation method under the framework of multi-task learning.
Compared with reference-based methods that evaluate based on the distance between the golden and generated response, the proposed RADE explicitly discriminates whether the reference or candidate response fits the dialogue context better.
To relieve the one-to-many problem, we augment RADE with a joint response generation task, which aims to perceive the range of feasible candidate responses. 
To improve the performance of RADE with the limited dataset, we propose a two-stage training strategy, including cross-domain pre-training and task-specific finetune.

\begin{figure*}
\centering
\includegraphics[width=1\textwidth]{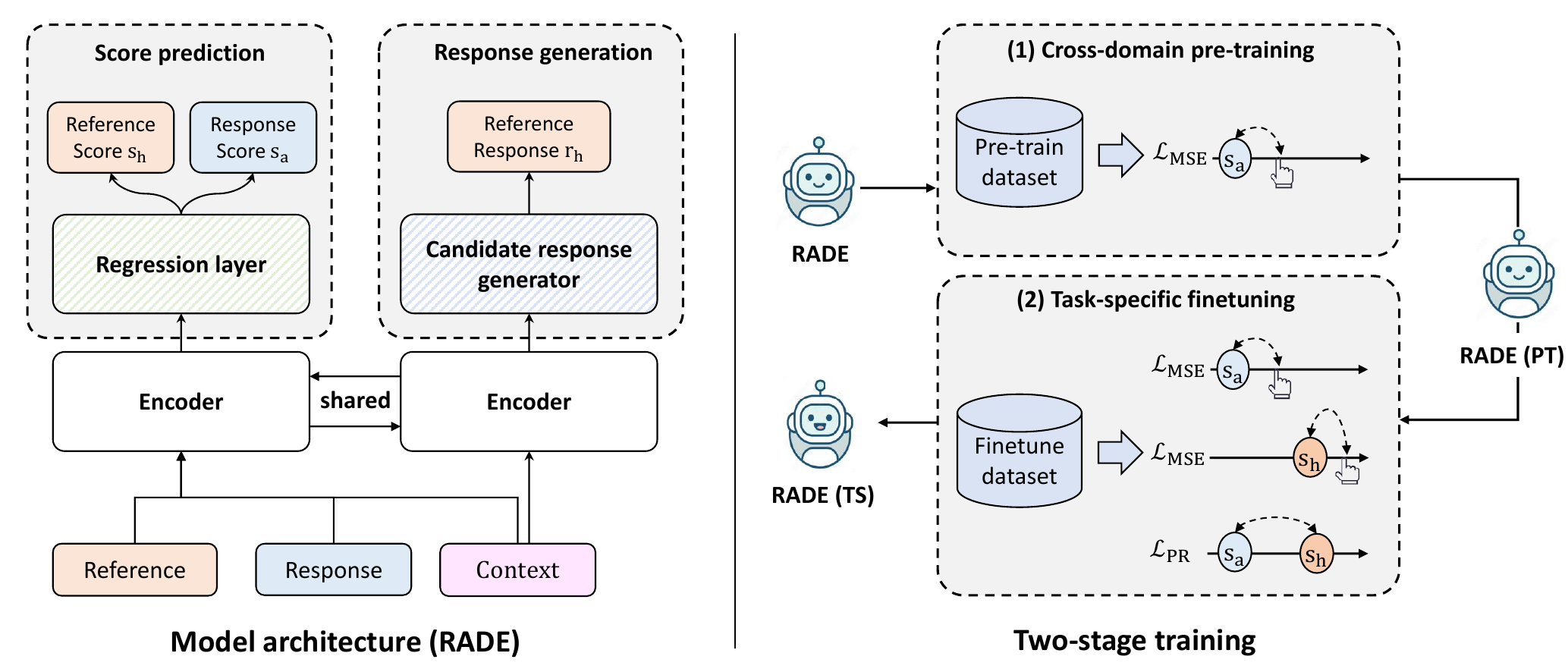}
    \caption{\textbf{Left:} An overview of our model which consists of an encoder, a regression layer, and a response generator. \textbf{Right:} Our two-stage training process with cross-domain \textbf{p}re-\textbf{t}raining (PT) and \textbf{t}ask-\textbf{s}pecific finetuning (TS).}
 \label{fig:model}
\end{figure*}

The architecture of RADE is illustrated in Figure~\ref{fig:model}, which comprises a posterior encoder, a regression layer, and a candidate response generator.

\paragraph{Posterior encoder.}
The posterior encoder encodes the dialogue context $c$, reference response $r_h$, and model-generated response $r_a$ into hidden representation.
In particular, we first concatenate $c$, $r_h$ and $r_a$ together into $X$ with a specific token \code{[SEP]}:
\begin{equation}
X=\{c~\text{\code{[SEP]}}~r_h~\text{\code{[SEP]}}~r_a\}
\end{equation}
Then the concatenated sequence is fed into a transformer-based encoder to get the representation $\mathbf{H}\in \mathbb{R}^{|X|\times d}$:
\begin{equation}\label{eq:encoder}
\mathbf{H}=\operatorname{Encoder}(X),
\end{equation}
where $d$ is the hidden size of encoder, $|X|$ is the length of sequence $X$.

\paragraph{Regression layer.}
The regression layer aggregates the representation $\mathbf{H}$ and predicts the scores of both reference and candidate response simultaneously.
Specifically, a pooling layer aggregates the token-level representation into a sequence-level representation: 
$\mathbf{h}\in \mathbb{R}^{d \times 1}$:
\begin{equation}
\mathbf{h}=\operatorname{Pooling}(\mathbf{H})
\end{equation}
Then, a feedforward network takes $\mathbf{h}$ as input to predict the score of both reference and candidate response:
\begin{equation}
(\hat{s_h},\hat{s_a}) =\operatorname{FeedForward}(\mathbf{h}),
\end{equation}
where $\hat{s_h}$ and $\hat{s_a}$ denote the predicted score of  $r_h$ and $r_a$, respectively.

\paragraph{Candidate response generator.}

To relieve the one-to-many problem, we devise a candidate response generator to perceive the range of feasible candidate responses~\citep{ZhangmingChan2021EnhancingTO}.
Specifically, a Transformer-based generator learns to generate reference responses autoregressively for a specific context.
We first encode the dialogue context $c$ using a encoder:
\begin{equation}
\hat{\mathbf{h}} = \operatorname{Encoder}{(c)},
\end{equation}
where the $\operatorname{Encoder}$ shares the same parameters with the posteriori encoder in Eq.~(\ref{eq:encoder}).
Then, we apply a Transformer-based decoder $\operatorname{Decoder}$ to model the generation probability of reference response $r_h$:
\begin{equation}\label{eq:generation}
    \begin{aligned}
        P(r_h|c)=\prod_{t=1}^T \operatorname{Decoder}(r_h^{(t)}|r_h^{(<t)},\hat{\mathbf{h}}),
    \end{aligned}
\end{equation}
where $T$ denotes the length of $r_h$.

Compared with the previous reference-free methods, which estimate the relation between context and response only with the knowledge acquired from their training data, RADE explicitly takes the pre-created response as a benchmark to reduce the data-induced bias when generalizing to agnostic scenarios.
Moreover, different from existing reference-based methods, which use the pre-created response as the golden standard without considering the semantic diversity of the response, we relieve the one-to-many problem via auxiliary response generation tasks.
The share encoder enhances the capability of context representation which augment the performance of score-predicting task through multi-task learning.

\subsection{Two-stage training}
The neural-based model has been proven prone to data-induced bias, but it is costly to annotate a large dataset in every specific task.
Therefore, we propose a two-stage strategy that includes: (1) \emph{cross-domain pre-training}, and (2) \emph{task-specific fine-tuning}, keeping a tradeoff of performance between in- and cross-domain.
As shown in Figure~\ref{fig:model} (right), we pre-train our model based on existing human-annotated datasets from different downstream tasks of open-domain dialogue to improve the generalizability~\citep{ZhengYe2021TowardsQD}.
Since the cross-domain datasets suffer from domain gaps and no pair-wised score, we finetune our model in the next stage with newly collected task-specific datasets.

\paragraph{Cross-domain pre-training.}
The pre-training datasets contain 54,438 dialogue-level examples collected from different downstream tasks, covering a wide range of domains (see more details in Table~\ref{pre-train-data}).
For learning the coarse-grain judgment of generated response without human-annotated reference scores, our model is first pre-trained  by minimizing a new cross-domain pre-training loss $\mathcal{L}_{\text{Cross}}$. 
Concretely, the $\mathcal{L}_{\text{Cross}}$ is composed of score-prediction loss and generation loss, which can be formulated as:
\begin{equation} 
\mathcal{L}_{\text{Cross}} = \mathcal{L}_{\text{MSE}}(\hat{s_a}, s_a) + \mathcal{L}_{\text{GEN}},
\end{equation}
where $\hat{s_a}$ and $s_a$ denote the human-annotated score and the predicted score of the candidate response and $\mathcal{L}_{\text{MSE}}(\hat{s_a}, s_a) = (\hat{s_a} - s_a)^2$.
$\mathcal{L}_{\text{GEN}}$ is the response generation loss, which is defined as:
\begin{equation} \label{eq:gen}
    \begin{aligned}
        \mathcal{L}_{\text{GEN}} &=-\log P(r_h|c),
    \end{aligned}
\end{equation}
where $P(r_h|c)$ is the generation probability of $r_h$ defined in Eq.~(\ref{eq:generation}).

\paragraph{Task-specific finetuning.}
We next finetune our model with newly annotated datasets to enhance the performance when evaluating task-specific dialogue agents. 
The optimize objective $\mathcal{L}_{\text{In}}$ is composed of score-prediction loss, generation loss, and pair-wised  ranking loss, which can be formulated as:
\begin{equation} 
\begin{split}
\mathcal{L}_{\text{In}}= & \mathcal{L}_{\text{MSE}}(\hat{s_a}, s_a) + \mathcal{L}_{\text{MSE}}(\hat{s_h}, s_h) + \\
& \mathcal{L}_{\text{GEN}}+\mathcal{L}_{\text{PR}}
\end{split}
\end{equation}
where $\mathcal{L}_{\text{MSE}}(\hat{s_a}, s_a)$ and $\mathcal{L}_{\text{MSE}}(\hat{s_h}, s_h)$ are MSE score-prediction loss of reference response and candidte response, respectively.
$\mathcal{L}_{\text{GEN}}$ is the generation loss as defined in Eq.~(\ref{eq:gen}).
$\mathcal{L}_{\text{PR}}$ is the pair-wise ranking loss defined as:
\begin{equation}
    \mathcal{L}_{\text{PR}}=-g(s_h,s_a)\log \frac{{\rm e}^{\hat{s_a}}}{{\rm e}^{\hat{s_h}} + {\rm e}^{\hat{s_a}}},
\end{equation}
in which $g(s_h,s_a)$ is a labeling function defined as:
\begin{equation}
    \begin{aligned}
    g(s_h,s_a)&=\begin{cases}
                    0,\quad s_h \geq s_a  \\
                    1,\quad s_h < s_a 
                    \end{cases}
    \end{aligned}
\end{equation}

The $\mathcal{L}_{\text{PR}}$ is introduced to assure that the rank order of the predicted scores satisfies the pre-annotated order.
Compared to reference-free models that inherently favor outputs from their underlying models or those trained on similar datasets, RADE is specifically optimized to align with human intentions and effectively alleviate this bias.

\section{Experimental Setup}\label{sec:exp}

\begin{table*}[!t]
\centering
\scalebox{0.8}{
\begin{tabular}{@{} l ccc ccc ccc  @{}}
\toprule
\multirow{2.5}{*}{\textbf{Methods}}
& \multicolumn{3}{c}{{\textbf{ChitChat}}} 
& \multicolumn{3}{c}{\textbf{{Empathetic Dialogue}}} 
& \multicolumn{3}{c}{{\textbf{PersonaChat}}} 
\\ 
\cmidrule(lr){2-4} \cmidrule(lr){5-7}   \cmidrule(lr){8-10} 

& $r$ & $\rho $ & $\tau$ 
& $r$ & $\rho $ & $\tau$ 
& $r$ & $\rho $ & $\tau$ 
\\ 

\hline

\rowcolor{Gainsboro}  \multicolumn{10}{l}{\textit{Reference-free methods}}\\
 FED$_{\text{E}}$~\citep{mehri2020unsupervised} 
 & 0.241 & 0.254 & 0.177 
 & 0.202 & 0.218 & 0.218 
 & 0.138 &0.120 &0.086  \\
 
 FED$_{\text{U}}$~\citep{mehri2020unsupervised} 
 & 0.235 & 0.248 &0.171 
 & 0.147 & 0.156 & 0.106 
 & 0.145 & 0.162 & 0.117 \\
 
 QuesEval~\citep{scialom-etal-2021-questeval}
 & 0.045 & 0.021 & 0.013
 & 0.069 & 0.084 & 0.057 
 & -0.003 & 0.034 & 0.0237    \\
 
 UniEval ~\citep{MingZhong2022TowardsAU} 
 & 0.456 & \underline{0.470} & \underline{0.312} 
 & 0.403 & 0.435 & 0.286 
 & \underline{0.306}  & \underline{0.338} & \underline{0.244}  \\

 DialoRPT~\citep{gao2020dialogrpt} 
 & -0.066$^*$ & -0.044$^*$ & -0.031$^*$ 
 & 0.267 & 0.244 & 0.166 
 & -0.077$^*$ & -0.069$^*$ & -0.049$^*$\\
 
 GRADE~\citep{LishanHuang2020GRADEAG}
 & \underline{0.491} & 0.434 &0.300 
 & \underline{0.549} & \underline{0.568} & \underline{0.398} 
 & -0.031$^*$ & -0.005 & -0.030$^*$ \\

QuantiDCE~\citep{quantdce}
 & 0.348 & 0.300 & 0.202
 & 0.498 & 0.507 & 0.351
 &  0.162 & 0.182 & 0.130 \\
 
\hline

\rowcolor{Gainsboro} \multicolumn{10}{l}{\textit{Reference-based lexicon-level methods} }  \\

ROUGE-L~\citep{lin-2004-rouge}
& 0.215 & 0.178 &0.129 
& 0.213 & 0.214 & 0.148 
& {0.118} & { 0.114} & {0.079}  \\

BLEU-2 ~\citep{papineni2002bleu}
& 0.201 & 0.200 & 0.158 
& 0.057 & 0.041$^*$ & 0.032 
& 0.060 & 0.039 & 0.031  \\

METEOR  ~\citep{banerjee2005meteor}
& 0.202 & 0.188 &0.129 
& 0.182 & 0.194 & 0.132 
& 0.099 & 0.051 & 0.035 \\ 

\hline
\rowcolor{Gainsboro} \multicolumn{10}{l}{\textit{Reference-based semantic-level methods} }  \\

 BERTScore ~\citep{TianyiZhang2019BERTScoreET} 
 & 0.296 & 0.243 & 0.213 
 & 0.167 & 0.243 & 0.173 
 & 0.278 & 0.292 & 0.196 \\
 
 BARTScore ~\citep{MichaelLewis2019BARTDS}
 & 0.133 & 0.057 & 0.039 
 & 0.256 & 0.253 & 0.173 
 & 0.143 & 0.168 & 0.115 \\
 
 RUBER~\citep{ChongyangTao2017RUBERAU} 
& 0.332 & 0.351 & \underline{0.369}
& 0.252 & 0.256 & 0.183 
&0.122 & 0.123 & 0.089 \\

 BLEURT~\citep{sellam2020bleurt} 
 & 0.353 & 0.363 & 0.249 
 & 0.343& 0.337& 0.232 
 & 0.105 & 0.140 & 0.102\\
 
BERT$_\text{MLP}$$^\dagger$~\citep{bert}
& 0.304 & 0.301 &0.192
& \underline{0.501} & \underline{0.537} & \underline{0.373} 
& \underline{0.331} & \underline{0.360} & \underline{0.251}  \\

BART$_\text{MLP}$ $^\dagger$~\citep{MichaelLewis2019BARTDS}
& \underline{0.431} & \underline{0.440} & 0.312 
& 0.412 & 0.447 & 0.356 
& 0.310 & 0.335 & 0.242  \\

\hline
\rowcolor{Gainsboro} \multicolumn{10}{l}{\textit{Reference-assisted methods}} \\

\textbf{RADE} (Pre-trained model, PT) 
& 0.472 & 0.491 & 0.334
& 0.650 & 0.601 & 0.427 
& 0.386 & 0.390 & 0.285\\ 

\textbf{RADE} (Task-specific model, TS) 
& \hllemon{\textbf{0.601}} & \hllemon{\textbf{0.569}} & \hllemon{\textbf{0.409}} 
& \hllemon{\textbf{0.863}} & \hllemon{\textbf{0.849}} & \hllemon{\textbf{0.685}} 
& \hllemon{\textbf{0.470}} & \hllemon{\textbf{0.465}} & \hllemon{\textbf{0.347}} 
 \\ 
\midrule

\hline
\rowcolor{Gainsboro} \multicolumn{10}{l}{\textit{Ablation Study}} \\
- w/o $\mathcal{L}_{\text{PR}}$
 & 0.503  & 0.514  & 0.353
 & 0.773  & 0.756  & 0.613    
 & 0.406 & 0.403  & 0.313 \\

- w/o $\mathcal{L}_{\text{GEN}}$ 
& 0.451  &    0.482 &   0.332 
& 0.751  &    0.740&   0.602
& 0.387 &    0.372 &   0.272  \\
\bottomrule
\end{tabular}
}
\caption{\textbf{Results on three benchmarks.}
The metrics $r$, $\rho $, and $\tau$ indicate the Pearson's $\rho$, Spearman's $r$, and Kendall'$\tau$. 
All values are statistically significant to p-value < 0.05 unless marked by $^*$. 
Methods with $^\dagger$ are implemented by ourselves.
We \underline{underline} the best results of each group of baselines methods and \textbf{bold} the best results of all methods.
The bottom of the table show the ablation study, where the proposed RADE is compared with several variants (-w/o: without). See section \ref{sec:ablation} for details.
}
\label{results}
\end{table*}

\subsection{Dataset and evaluation metrics}
We mainly conduct experiments on the three datasets annotated in Section~\ref{sec:ref-human}. 
We further evaluate the models on two existing benchmarks, USR-TopicChat and USR-PersonaChat~\citep{mehri2020usr}, to examine the generalizability of our method. 
The evaluation metrics include Pearson ($r$), Spearman ($\rho$), and Kendall ($\tau$) correlation, which measures the linear relationship, monotonic relationship, and the ordinal association between automatic evaluation and human evaluation, respectively\footnote{We use SciPy (\url{https://scipy.org/}) to calculate the scores.}.
We abbreviate the Pearson, Spearman, and Kendall correlation as $r$, $\rho $, and $\tau$ for simplicity.

\subsection{Implementation details}
We initialize the parameters of the encoder and decoder with BART~\cite{MichaelLewis2019BARTDS}, a Transformer-based pre-trained model.
BART is well-suited to our proposed model because it is capable of both text representation tasks and text generation tasks.
We optimize the model using Adam optimizer with parameters $\beta_1=0.98$, $\beta_2=0.97$, and the learning rate of $5e{-}5$.
The model is trained up to 10 epochs, and we tune the hyper-parameters and pick the checkpoint on the development set.
The training of the model can be done within 5 hours using two 2080Ti GPUs.
We denote the RADE model that pre-trained on cross-domain datasets as \textbf{RADE (PT)}, and the model that further finetuned on task-specific data as \textbf{RADE (TS)}.

\subsection{Baselines}
We compare our method with two types of baselines: reference-based and reference-free methods.

The reference-free baselines include: 
\emph{DialoRPT}~\citep{gao-etal-2020-dialogue}, which trained on large-scale social media feedback data to predict ranking-based scores;
\emph{GRADE}~\citep{LishanHuang2020GRADEAG}, which enhances the contextualized representations via topic-level commonsense graphs and predicts the score using a regression module;
\emph{FED}~\cite{ShikibMehri2020UnsupervisedEO}, an unsupervised dialogue evaluation model based on DialogGPT;  \emph{UniEval}~\citep{MingZhong2022TowardsAU}, which evaluates the response from multiple perspectives;
\emph{QuesEval}~\citep{scialom-etal-2021-questeval}, which evaluates the fact-based text using summarizing asks.
  
The reference-based baselines include:  \emph{RUBER}~\citep{ChongyangTao2017RUBERAU}, an unsupervised evaluation metric considering the similarity of the response with dialog context and reference;
\emph{BERTScore}~\citep{TianyiZhang2019BERTScoreET}, which employs BERT to greedily match the response and the ground truth at the token level;
\emph{BLEURT}~\citep{sellam2020bleurt}, which is a BERT-based model pre-trained with millions of synthetic examples;
\emph{BARTScore}~\citep{de2020bart}, which weights the log-likelihood of the generated response as the score.
We also test three reference-based lexical-level metrics: \emph{ROUGE-L}, \emph{BLEU-2}, and \emph{METEOR}.

Moreover, we implement two reference-based baselines, BERT$_\text{MLP}$ and BART$_\text{MLP}$, which are trained with the same human-annotated datasets as RADE, and provide a reasonable comparison with our proposed model.
Specifically, we obtain the text representations of the dialogue using BERT or BART and then feed the representations into a multi-layer perception to calculate the scores.
For a more comprehensive analysis, we also fine-tune the two strongest baselines, QuantiDCE and GRADE, on our cross-domain datasets as well as our self-collected datasets, respectively.

\section{Results and Analysis}\label{sec:ra}

\subsection{Experimental results}
\paragraph{Overall performance.}
Table~\ref{results} shows the experimental performance for all methods.
Overall, RADE achieves the best performance in three benchmarks in terms of all metrics.
Concretely, the pre-trained model RADE (PT) gets better or comparable correlation with human judgment than the best baseline method on three dialogue tasks.
The task-specific model RADE (TS), fine-tuned with the newly collected reference-assisted data, establishes a new state-of-the-art by improving the performance by about 30\% on average compared to RADE (PT).
For example, RADE (TS) gets $r=0.601$, $\rho=0.569$ in the ChitChat domain, and pushes $r$ to $0.863$ ($0.314$ absolute improvements), $\tau$ to $0.685$ ($0.287$ absolute improvements) in EmpaDial domain.
This result suggests that training with in-domain datasets is critical to enhancing the task-specific evaluation capability of RADE.
For a more comprehensive comparison, we also train the two strongest baselines (QuantiDCE and GRADE) with our cross-domain and self-collected datasets, respectively.
And the result and analysis are provided in Appendix~\ref{app:fine-tune-comparison}.

\paragraph{Generalizability.}
We find that the performance of the reference-free method varies dramatically across domains.
For example, GRADE and QuantiDCE, trained in the chitchat domain, achieve high correlations with human judgment in ChitChat and EmpaDial but perform poorly in PersonaChat.
The result indicates that the contextual representation capabilities of unsupervised methods are limited by their training data and, therefore, are prone to data-induced bias, decreasing their performance when employing agnostic scenarios.
In contrast, the gap between the proposed RADE (PT) methods across different domains is relatively small.
These results indicate that RADE has better generalizability than reference-free methods due to the assistance of reference and the proposed cross-domain training strategy.

\paragraph{Results on USR benchmarks.}
We further examine our methods on two USR datasets~\citep{mehri2020usr} to verify the efficiency and robustness of RADE when generalizing to existing dialogue evaluation benchmarks.
The results are listed in Table~\ref{usr_eval}.
Experiments show that RADE, which has not explicitly trained on these datasets, achieves better or comparable results to previous supervised methods.
See Appendix~\ref{app:usr} for more results and details.

\begin{table}[!ht]
\centering\small
\setlength\tabcolsep{12pt}

\begin{tabular}{@{} m{1.3cm} cc cc  @{}}
\toprule
\multirow{2.5}{*}{\textbf{Methods}}
& \multicolumn{2}{c}{USR-Topical} 
& \multicolumn{2}{c}{USR-Pearsona} 
\\
\cmidrule(lr){2-3} \cmidrule(lr){4-5} 

& $r$ & $\rho$ & $r$ & $\rho$ \\

\midrule

GRADE
& 0.200 & 0.217 & 0.358 & 0.352 \\

USR
& \underline{0.412} & \underline{0.423} & 0.440 & 0.418  \\ 

USL-H
& 0.322 & 0.340 & \textbf{0.495}
& \textbf{0.523} \\ 

\midrule

METEOR 
& 0.336 & 0.391 & 0.253 & 0.271 \\

BERTScore
& 0.298 & 0.325 & 0.152 & 0.122 \\

BLEURT
& 0.216 & 0.261 & 0.065 & 0.054 \\

\midrule

\textbf{Ours} 
&  \hllemon{\textbf{0.480}}& \hllemon{\textbf{0.466}}  &\hllemon{\underline{ 0.451}} &  \hllemon{\underline{0.465}} \\ 

\bottomrule
\end{tabular}
\caption{Results on USR-TopicalChat and USR-PearsonaChat~\citep{mehri2020usr}.
}
\label{usr_eval}
\end{table}

\subsection{Ablation study}\label{sec:ablation}
We perform an ablation study to investigate the influence of different components in our methods.
We examine two ablative variants:
(1) w/o $\mathcal{L}_{\text{PR}}$: we remove the ranking-based loss $\mathcal{L}_{\text{PR}}$ to verify its effectiveness (w/o $\mathcal{L}_{\text{PR}}$);
(2) w/o $\mathcal{L}_{\text{GEN}}$: we remove the $\mathcal{L}_{\text{GEN}}$ to verify training with response generation task jointly can improve the predicting correlation with human judgment.

Table \ref{results} presents the results.
Overall, the variants of our methods show a decreased performance compared to the base model.
For example, Pearson drops 0.10, 0.09, and 0.07 in three benchmarks, respectively, after the $\mathcal{L}_{\text{PR}}$ is removed.
This result indicates that ranking-based loss can enhance performance by explicitly building the relation between response and reference.
After removing the $\mathcal{L}_{\text{GEN}}$, the correlation in all benchmarks has a prominent decrease, e.g., Spearman correlation drops by 0.15, 0.10, and 0.09, respectively.
The results suggest that the auxiliary response generation task improves the representation capability of our method and relieves the one-to-many problem.

\subsection{Case study}
Our case studies demonstrate that RADE is more consistent with human judgment than baselines. Details about our case studies are available in Appendix \ref{app:case}.

 \begin{figure}[t]
    \centering
\includegraphics[width=0.5\textwidth]{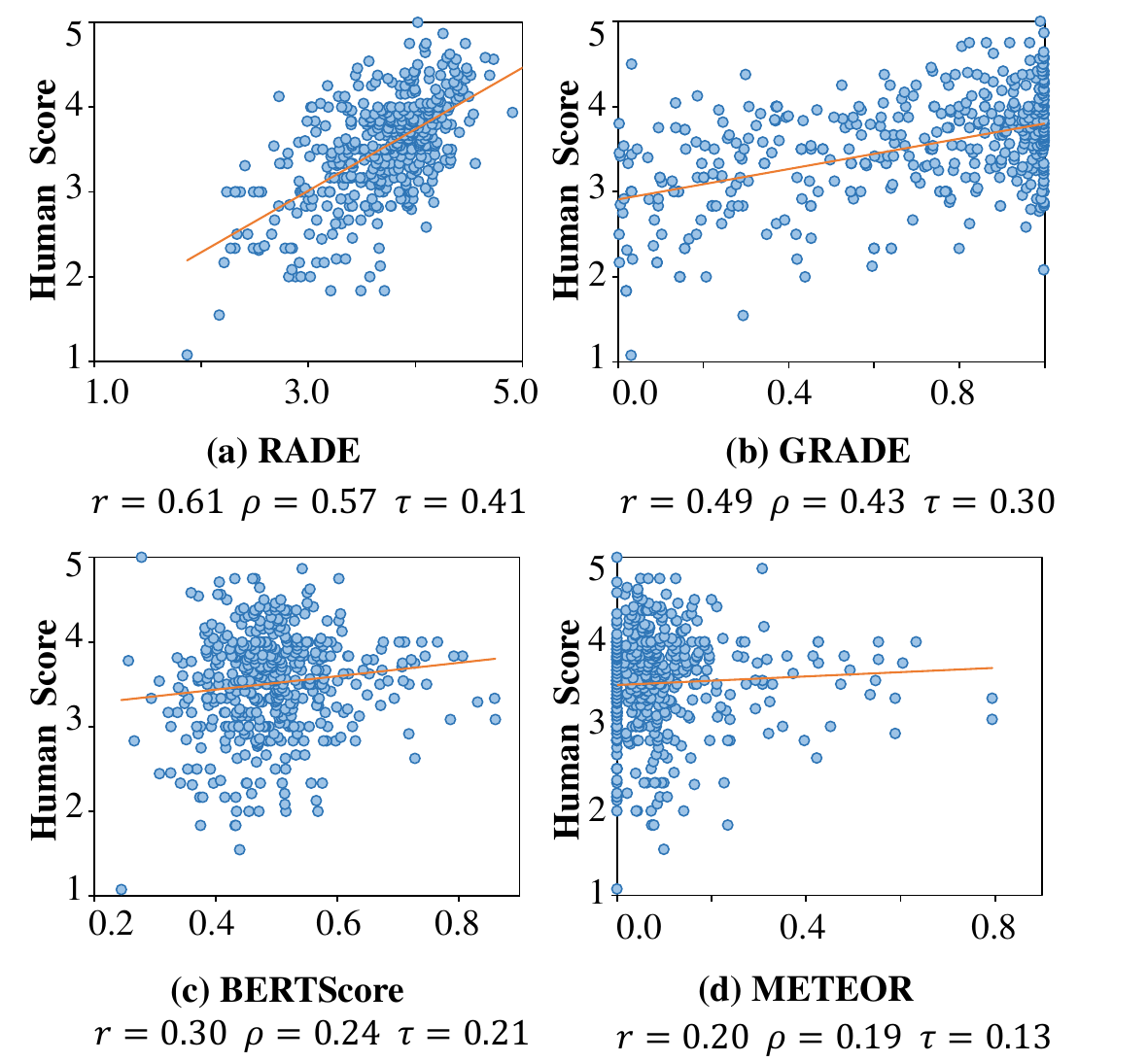}
    \caption{Score correlation of automatic evaluation and human evaluation on the EmpaDial domain. The horizontal axis indicates the different automatic evaluation methods, and the vertical axis indicates human rating.
    }
\label{correlation}
\end{figure}

\subsection{Qualitative analysis}
To explain more intuitively, we show the scatter plots against human judgments for different automatic evaluation methods (i.e., RADE, GRADE, BERTScore, METEOR) on the EmpaDial dataset in Figure~\ref{correlation}.
As shown in Figure~\ref{correlation}~(a), our method RADE achieves a stronger correlation with human judgment than the other methods.
Figure~\ref{correlation}~(d) illustrates that METEOR scores are zero or extremely low for the most response.
It results from the one-to-many nature of open-domain dialogue, and word overlapping occasionally occurs.
Figure~\ref{correlation}~(c) suggests that the BERTScore scores are mainly concentrated in the range of 0.3-0.6, indicating no significant differentiation between the different responses.
Figure~\ref{correlation} (b) shows that GRADE achieves a better correlation with human judgments.
However, the distribution of GRADE predicted scores is concentrated in the high-scoring band, resulting in a low distinction of responses; 
RADE uses reference as a benchmark and thus has a more balanced distribution of predicted scores.

\section{Discussions}

\paragraph{The impact of the training data scale.}
To explore the minimum data scale required for our method, we train RADE using different amounts of randomly sampled annotated data. 
We observe a minor degradation in RADE's performance as the amount of data decreases. For example, when training on 2,400 examples from the EmpatheticDialogue dataset, RADE(TS) achieves Pearman'r=0.837 and Spearman'rho=0.829; whereas with 1,200 examples, it obtains Pearman'r=0.807 and Spearman'rho=0.806. 
All results are averaged over three runs. Moreover, we find that RADE outperforms all baselines with only 800 training examples in three datasets, respectively.

\paragraph{The difference between golden and candidate Responses.}
\emph{Golden response} refers to a scenario where there is only one correct response, and any different response is given a low score. For example, BERTScore calculates the cosine similarity between the golden and model-generated response. However, \emph{Candidate responses} implies that there can be multiple correct answers, which is more flexible and human-intuitive. And RADE is optimized to align with this human intention using generative and pairwise-ranking loss. If more references are available, the RADE can consider multiple valid responses to make more reliable evaluations. 
To achieve this, we can concatenate model-generated responses with different references. However, due to the limitation of our datasets, we concatenate one reference and model-generated response, which are then fed to the encoder.

\paragraph{Employing RADE when the reference response is not available.}
Considering the reference is not always available in real-world scenarios, we design two alternatives to enable RADE, i.e., constructing a pseudo-reference via retrieval or generative method. 
We verify the two solutions on the FED dataset and the details can be found in Appendix~\ref{app:fed}.

\section{Conclusion}

We have presented a new reference-assist dialogue evaluation (RADE) method to address the one-to-many problem when evaluating open-domain dialogue systems. RADE evaluates the response generated by open-domain dialogue agents with the assistance of reference response.
In addition, we have curated the reference-assisted dialogue evaluation datasets by expanding three existing datasets via a pairwise human annotation. The extended datasets contain over 10K dialogues.
Extensive experiments on three extended datasets and two existing benchmarks have verified the effectiveness and robustness of the proposed methods and their generalizability.

\section*{Limitations}
The main limitation of this paper is the need for human-labeled reference responses. 
We will explore automated or human-machine collaboration methods to reduce the cost of annotation in the next stage.
Another limitation is that we need to explore whether other auxiliary tasks can also enhance the performance of score prediction.
In the future,  we also plan to reproduce the proposed method for other, less resource-rich languages.

\section*{Ethics Statement}
The paper proposes a dialogue evaluation method, which is intended to evaluate open-ended dialogue on topics such as books and movies. 
A new dataset is developed using some existing dialogue systems, such as DialoGPT, which are trained on large-scale web data that is known to contain biased or discriminatory content.
The datasets that we trained on may also include subjective knowledge (comments on movies) that may express the bias of the writers.


\newpage
\bibliography{reference}
\bibliographystyle{acl_natbib}
\clearpage

\newpage
\appendix
\section{Appendix}
\subsection{Human Evaluation Details}
\subsubsection{Details for Data Preparation}\label{models}
We first employ the generation models to generate one more response for our human annotation proposed in Section ~\ref{sec:task}.
The annotators are instructed to rate the newly generated responses.
Specifically, we employ the following generation model:

\begin{itemize}
    \item \textbf{Blenderbot}~\citep{blenderbot}: Blender is a conversational agent based on the large-scale model that mainly focuses on generating personal,  engaging, knowledgeable, and empathetic responses.
    \item \textbf{DialogGPT}~\citep{zhang2019dialogpt}: DialogGPT is a  large, tunable neural conversational response generation model.
    \item \textbf{KEMP}~\citep{QintongLi2023KnowledgeBF}: KEMP is an emotional dialogue agent enhanced with a knowledge-enriched context graph.
    \item \textbf{MoEL}~\citep{ZhaojiangLin2019MoELMO}: MoEL is an emotional dialogue agent based on encoder-decoder architecture. MoEL softly combines the response representation from different decoders, each focusing on one type of emotion.
    \item \textbf{MIME}~\citep{majumder2020mime}: MIME is an empathetic dialogue model considering polarity-based emotion clusters and emotional mimicry.
    \item \textbf{EmpDG}~\citep{QintongLi2019EmpDGMI}:  EmpDG is a multi-resolution empathetic chatbot enhanced by exploiting user feedback.
    \item \textbf{PersonaGPT}~\citep{tang2021persona}: PersonaGPT is a GPT2-based open-domain dialogue agent designed to generate personalized responses.  
\end{itemize}
\begin{table}[htbp]
\small
\centering
\begin{tabular}{@{}lccc@{}}
\toprule
Model & DSTC & EmpaDial & PersonaChat \\ 
\midrule
Blenderbot   &  812    &         &        500     \\
DialoGPT      &    1278     &        &    500     \\
KEMP   &      &      3014   &             \\
MoEL      &      &      231    &             \\
MIME      &      &   242    &             \\
EmpDG      &      &   535    &             \\
PersonaGPT      &      &          &     3000     \\ 
\bottomrule
\end{tabular}
\caption{The data distribution of seven well-performing dialogue models, which are used for extend corresponding dataset.}
\label{prepare}
\end{table}

As shown in Table ~\ref{prepare}, we extend the DSTC dataset with \emph{Blenderbot} and \emph{DialoGPT}, the Empathetic Dialogue dataset with \emph{KEMP}, \emph{MoEL}, \emph{MIME} and \emph{EmpDG}; the Persona-Chat dataset with \emph{Blenderbot}  and \emph{PersonaGPT}.

Since ~\citeauthor{blenderbot} points out the length of the utterances is crucial to human judgments, i.e., too short responses are seen as dull, we only sample the example with at least two turn interactions with an average length of utterance no more than 25 vocab.
And we randomly split the train-dev-test of collected datasets as Chitchat (1490/300/300, 5/1/1), Empathetic Dialogue (3022/500/500, 6/1/1), Persona Chat (3000/500/500, 6/1/1).

\subsubsection{Annotation Guideline}\label{ui}

Table ~\ref{guide} provides detailed instructions for the annotators to s
help them understand the setting of our annotation task.
\begin{table}[htbp]
\small
\centering

\begin{tabular}{@{}m{7.5cm}@{}}
\toprule
\multicolumn{1}{c}{\textbf{Annotation Guideline}}
 \\
\hline
\rowcolor{Gainsboro}  \multicolumn{1}{l}{\textit{ Instruction}}\\ 
\hline
You need to read the context for each conversation to understand the specific context. Afterward, compare the two responses and determine which is better on the given metric. 
Since we have given a score to the reference response, you should take it as the benchmark and rate the generated response.  
\\
\hline
\rowcolor{Gainsboro}  \multicolumn{1}{l}{\textit{ Dataset}} 
\\
\hline
(1) context: The historical interaction between two partners.
\\
(2) (reference,$s_h$): The reference response and corresponding score. 
\\
(3) response: The  response generated via agent which you need to rate.
\\
\hline
\rowcolor{Gainsboro}  \multicolumn{1}{l}{\textit{Rating Details}}
\\
(1) If the  generated responds is better, the scores you give should be more than $s_h$. 
\\
(2) If the  generated responds is worse, the scores you give should be less than $s_h$. 
\\
(3) If there is no significant difference between the two response, you can give the same score as $s_h$.
\\
\bottomrule
\end{tabular}
\caption{The  guideline used for our human annotation.}
\label{guide}
\end{table}

\subsubsection{User Study}\label{user_study}
The dialogue can be evaluated from multiple perspectives.
Some perspectives are universal to assess all dialogue agents, e.g., fluency, and relevance, while the other metrics are only used for task-specific dialogue agents.
\begin{figure}[htbp] 
\centering
    \includegraphics[width=0.5\textwidth]{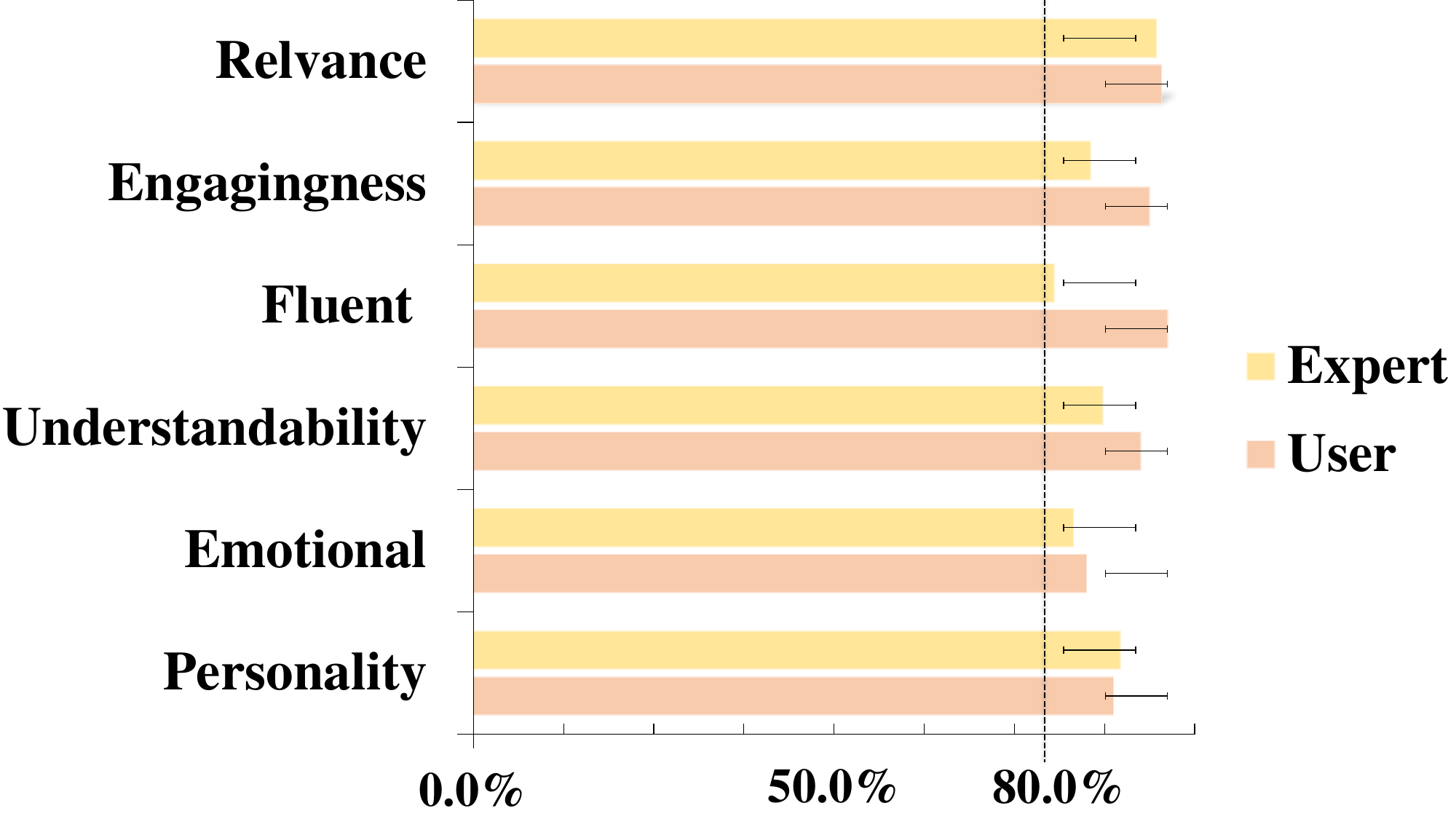}
    \caption{Result  of two-role user study. }
    \label{fig:user_study}
\end{figure}
For example, the emotion-aware is a critical property for empathetic dialogue but is less important for persona dialogue.
Therefore, we first simplify by sorting the possible aspects into two categories, i.e., the general view and the task-specific view.
The former contains relevance, engagingness, and fluency, while the latter consists of understandability, emotion-aware, and personality-aware, which correspond to chitchat dialogue, emotional dialogue, and persona dialogue.
To understand the relation between sub-metrics and overall quality, we conduct a user study to learn their preference for different sub-metrics.
Specifically, we invite 20 experts and 80 users, each of whom is asked to select the four most important ones from the sub-metrics.
The results are listed in Figure~\ref{fig:user_study}. 
The approval rates reflect the user preference for different sub-metrics, which can be used as a weight to calculate the overall score.
Moreover, we apply the softmax function on these weights to make them more interpretable.

\subsection{ Experiment Details}\label{app:details}

\subsubsection{Datasets for Pre-train Stage}\label{app:pre-train-data}
Our training process includes two stages, e.g., cross-domain pre-train and task-specific finetune.
We first pre-train the model on diverse open-domain dialogue datasets as listed in Table~\ref{pre-train-data} with the objective $\mathcal{L}_{\text{cross}}$.
The next stage relies on task-specific dataset with the  objective $\mathcal{L}_{\text{in}}$   
(see in section \ref{sec:ref-auto}).

These datasets are collected from \url{https://github.com/e0397123/dstc10_metric_track}, which contain a variety of open-domain dialogue, such as emotional dialogue, personalized dialogue, knowledge-grounded dialogue, and chitchat.
Every example in the datasets contains the dialogue \emph{context}, \emph{response} generated by dialogue agent, pre-created \emph{reference} response, and the \emph{score} of generated response which has been annotated for at least three people from several perspectives.
We use cross-domain datasets for pre-training to improve the robustness and generalisability of the models across different evaluation scenarios.

\begin{table}[ht]
\centering
\small
\caption{Statistics of our datasets used for pre-train stage.
AVG. Utts: the average of utterances per dialogue;
AVG. Words : the average of words per dialogue.}
\label{pre-train-data}
\begin{tabular}{@{} lrccc @{}}
\toprule
Dataset & Dialogue & AVG. Utts & AsVG. Words \\
\midrule
DSTC6-Eval    & 33,795 & 2.63 & 11.36  \\ 
DSTC7-Eval  & 9,711  & 3.83 & 13.40  \\ 
DSTC10-Eval & 9,291  & 4.00 & 14.15  \\ 
JSALT-Eval  & 741 & 3.47& 17.12\\
PersonaChat-Zhao & 900  & 5.13 & 11.77\\ 
\bottomrule
\end{tabular}
\end{table}

\subsubsection{Experimental Details on Our Benchmarks}
We show the details of our automatic evaluation experiments in Table \ref{automatic_all_eval}.
The BERTScore and BLEURT are computed based on the large version of Roberta.
As in Section~\ref{sec:exp}, we implement two reference-based baselines, BERT$_{\text{MLP}}$ and BART$_{\text{MLP}}$, using the same human-annotated datasets as RADE for training, and provide a reasonable comparison with our proposed model.
Specifically, the BERT$_\text{MLP}$ is built on the base version of BERT~\citep{bert}, while the BART$_\text{MLP}$ is built on the base version of BART~\citep{MichaelLewis2019BARTDS}.

\subsubsection{More Fair Comparison after Training}\label{app:fine-tune-comparison}
For a fair analysis, we pre-train the two strongest baselines (QuantiDCE and GRADE) with our cross-domain dataset. GRADE achieves Pearman'r=0.383, 0.378, -0.122, and QuantiDCE achieves  Pearman'r=0.408, 0.522, 0.238 in the ChitChat, EmpatheticDialogue, and Personachat datasets. However, our proposed RADE(PT) remains the best results (Pearman'r=0.472, 0.650, 0.386). 
We further fine-tune GRADE and QuantiDCE with our self-collected datasets for a more comprehensive analysis. GRADE achieves Pearman'r=0.413, 0.430, -0.013, and QuantiDCE achieves Pearman'r=0.458, 0.589, 0.278 in three datasets,  underperforming the proposed RADE(TS) (Pearman'r=0.601, 0.863, 0.470).

We skip pre-training/fine-tuning four baselines for the following reasons:
(1) UniEval and QuestionEval have been pre-trained on multiple datasets across various domains.  (2) The FED metric is unsupervised (cf. Shikib Mehri et al.)
(3) The DialoRPT has been trained on a sizeable human-feedback dataset (133M) covering various domains.
These analyses validate the superiority of our method. 

\begin{table*}[!ht]
\centering
\scalebox{0.8}{
\begin{tabular}{@{} m{7cm} cc cc cc @{}}
\toprule
\multirow{2.5}{*}{\textbf{Methods}}
& \multicolumn{2}{c}{USR-TopicalChat} 
& \multicolumn{2}{c}{USR-Pearsonachat} 
& \multicolumn{2}{c}{DailyDialogue} 
\\

\cmidrule(lr){2-3} \cmidrule(lr){4-5} \cmidrule(lr){6-7} 

& Pearson'$r$ & Spearman'$\rho$ 
& Pearson'$r$ & Spearman'$\rho$ 
&  Pearson'$r$ & Spearman'$\rho$ \\
\hline
\rowcolor{Gainsboro}\multicolumn{7}{l}{\emph{Reference-free methods}} \\
MAUDE~\citep{maude} 
& 0.044* & 0.083* 
& 0.345 & 0.298 
& -0.036* & -0.073*
\\

FED~\citep{mehri2020unsupervised} 
& -0.124 & -0.135 
& -0.028* & -0.000*
& -0.080* & 0.064*
\\

HolisticEval~\citep{he} 
& -0.147 & -0.123 
& 0.087* & 0.113* 
& 0.025*& 0.020*
\\

FlowScore~\citep{flowscore}
& 0.095* & 0.082*
& 0.118* & 0.079*  
& - & -
\\

QuestEval~\citep{scialom-etal-2021-questeval}
& 0.300 & 0.338 
& 0.176 & 0.236   
& 0.020* &0.006*
\\

USR~\citep{mehri2020usr} 
& \underline{0.412} & \underline{0.423} 
& 0.440 & 0.418   
& 0.057* & 0.057*
\\ 

GRADE~\citep{LishanHuang2020GRADEAG}
& 0.200 & 0.217
& 0.358 & 0.352 
& 0.278 & 0.253
\\

PredictiveEngage~\citep{pe} 
& 0.222 & 0.310 
& -0.003* & 0.033* 
& -0.133* & -0.135
\\

DialogRPT~\citep{gao2020dialogrpt} 
& 0.120 & 0.105*
& -0.064* & -0.083*  
& -0.000* & 0.037*
\\

DynaEval~\citep{dynaeval}
& -0.032* & -0.022*
& 0.149 & 0.171    
& 0.108* & 0.120*
\\

DEB~\citep{deb} 
& 0.180 & 0.116 
& 0.291 & 0.373   
& \underline{0.337} & \underline{0.363}  
\\

USL-H~\citep{mehri2020usr} 
& 0.322 & 0.340 & \underline{\textbf{0.495}} 
& \underline{\textbf{0.523}} 
& 0.108* & 0.093*
\\ 

\hline
\rowcolor{Gainsboro}\multicolumn{7}{l}{\emph{Reference-based lexicon-level methods}} \\

BLEU-4~\citep{papineni2002bleu} 
& 0.216 & 0.296
& 0.135 & 0.090* 
& 0.075* & 0.184
\\

METEOR~\citep{banerjee2005meteor} 
& \underline{0.336} & \underline{0.391} 
& \underline{0.253} & \underline{0.271}
& 0.093* & 0.010*
\\

ROUGE-L~\citep{lin-2004-rouge} 
& 0.275 & 0.287 
& 0.066* & 0.038* 
& \underline{0.154} & \underline{0.147} 
\\

\hline

\rowcolor{Gainsboro}\multicolumn{7}{l}{\emph{Reference-based semantic-level methods}}  \\

RUBER~\citep{ChongyangTao2017RUBERAU}
& 0.247 & 0.259 
& 0.131 & 0.190 
& -0.084* & -0.094* 
\\

BERT-RUBER~\citep{ChongyangTao2017RUBERAU} 
& \underline{{0.342}} & \underline{0.348}
& 0.266 & 0.248
& 0.134 &0.128
\\

BERTScore~\citep{TianyiZhang2019BERTScoreET}
& 0.298 & 0.325 
& 0.152 & 0.122* 
& 0.129 & 0.100*
\\

Deep AM-FM~\citep{am-fm}
& 0.285 & 0.268 
& 0.228 & 0.219 
& 0.026*& 0.022*
\\

ADEM~\citep{RyanLowe2017TowardsAA}
& -0.060* & -0.061* 
& -0.141 & -0.085*   
& 0.064*& 0.071*
\\

BLEURT~\citep{sellam2020bleurt} 
& 0.216 & 0.261
& 0.065* & 0.054*  
& \underline{0.176} & \underline{0.133}
\\

PONE~\citep{pone}
& 0.271 & 0.274 
& \underline{0.373} & \underline{0.375} 
& 0.163 & 0.163
\\

\hline
\rowcolor{Gainsboro}\multicolumn{7}{l}{\textit{Reference-assist}} \\
\textbf{Ours} (Pretrain-train model, PT)   
&  \hllemon{\textbf{0.480 }}& \hllemon{\textbf{0.466}}  
&\hllemon{{ 0.451}} &  \hllemon{{0.465}}   
& \textbf{0.356 } & \textbf{0.370}
\\ 

\bottomrule
\end{tabular}
}
\caption{
\textbf{Results on USR-TopicalChat, USR-PearsonaChat and Grade-DailyDialogue.}
We divide the methods in Reference-free, Reference-based and REDE, while the reference-based methods including semantic-level and lexicon-level.
The metrics $r$, $\rho $, and $\tau$ indicate the Pearson's $\rho$, Spearman's $r$, and Kendall'$\tau$. 
All values are statistically significant to p-value < 0.05 unless marked by $^*$. 
We \underline{underline} the best results of each group of baselines methods and \textbf{bold} the best results of all methods.
}
\label{full_usr_eval}
\end{table*}

\subsubsection{Results on  Existing Benchmarks}\label{app:usr}
We further examine three existing benchmarks, i.e., USR-TopicalChat, USR-PersonaChat and Grade-DailyDialogue to verify the efficiency and robustness of RADE when generalizing to agnostic scenarios.
USR-TopicalChat and USR-PersonaChat datasets are collected to assess dialog evaluation metrics, with examples containing the dialogue \emph{context}, \emph{reference}, \emph{response} and corresponding \emph{scores}, which three people have annotated.
The Grade-DailyDialogue contains high-quality open-domain conversations about daily life including diverse topics.
And the results are summarized in Table ~\ref{full_usr_eval}.

The experimental results show that RADE outperforms the state-of-the-art reference-free and reference-based methods on the USR-TopicalChat dataset.
For example, we push the Pearson correlation to 48.0\% (7\% definite improvement) and Spearman correlation to 46.6\% (4\% absolute improvement).
Moreover, RADE shows a stronger correlation with human judgment than existing reference-based methods on the second dataset.
 It achieves comparable, even better results with the reference-free methods except for USL-H.
The results demonstrate that our pre-trained model is more robust even under agnostic scenarios.

We also compare the two existing methods, and the results suggest a similar phenomenon as ~\ref{results}. 
Firstly, the reference-free methods achieve better consistency than reference-based methods, i.e., the former has the highest result of $r=41.2\%$, $\rho=42.3\%$ while the latter gets $r=34.2\%$, $\rho=34.8\%$  on the USR-TopicalChat dataset.
However, the reference-free methods suffer from more significant variance.
For example, the MAUDE gets $r=0.345\%$ and $\rho=0.298\%$  on the USR-PearsonChat dataset but gets $r=0.044\%$ and $\rho=0.083\%$  on the USR-TopicChat dataset.
It indicates that reference-free methods are more vulnerable and prone to data-induced bias.

\subsubsection{Case Study}\label{app:case}
To explain more intuitively, we show examples of automatic evaluation and them with human judgment in Table ~\ref{case:1}, ~\ref{case:2}, ~\ref{case:3}, suggesting that the scores of our methods are closer to human ratings.

\subsection{Presudo reference}\label{app:fed}

Since the original FED does not provide the reference response, we construct a pseudo-reference via retrieval or generative method. The former retrieves reference from a curated response corpus based on our cross-domain datasets via BM25 with the dialogue context as the query. The latter generates via a large language model GPT-3 based on the dialogue context. The results show that RADE(PT) obtains Pearman'r=0.381 and Spearman'rho=0.368 with the retrieved reference while achieving Pearman'r=0.343, Spearman'rho=0.347 with generative reference, outperforming the state-of-the-art baseline (QuantiDCE, Pearman'r=0.319, Spearman'rho=0.323). 

To further validate the generalizability of our method, we evaluate our proposed RADE(PT) on another challenging benchmark,  GRADE-Dailydialogue.
Our RADE(PT) achieves Pearman'r=0.356 and  Spearman'rho=0.370  with 5\%  and 2\% relative improvements compared to state-of-the-art baseline, indicating that our method can generalize to more challenging benchmarks.

\begin{table*}[htbp]
\centering

\scalebox{0.8}{
\begin{tabular}{@{} m{5.7cm} ccc ccc ccc @{}}
\toprule
\multirow{2.5}{*}{\textbf{Methods}}
& \multicolumn{3}{c}{{\textbf{ChitChat}}} 
& \multicolumn{3}{c}{\textbf{{Empathetic Dialogue}}} 
& \multicolumn{3}{c}{{\textbf{PersonaChat}}} 
\\ 
\cmidrule(lr){2-4} \cmidrule(lr){5-7}   \cmidrule(lr){8-10} 

& $r$ & $\rho $ & $\tau$ 
& $r$ & $\rho $ & $\tau$ 
& $r$ & $\rho $ & $\tau$ 
\\ 

\hline

\rowcolor{Gainsboro}  \multicolumn{10}{l}{\textit{Reference-free methods}}\\
 FED$_{\text{E}}$~\citep{mehri2020unsupervised} 
 & 0.241 & 0.254 & 0.177 
 & 0.202 & 0.218 & 0.218 
 & 0.138 &0.120 &0.086 \\
 
 FED$_{\text{U}}$~\citep{mehri2020unsupervised} 
 & 0.235 & 0.248 &0.171 
 & 0.147 & 0.156 & 0.106 
 & 0.145 & 0.162 & 0.117 \\
 
 QuesEval~\citep{scialom-etal-2021-questeval}
 & 0.045 & 0.021 & 0.013
 & 0.069 & 0.084 & 0.057 
 & -0.003 & 0.034 & 0.0237   \\
 
 UniEval ~\citep{MingZhong2022TowardsAU} 
 & 0.456 & \underline{0.470} & \underline{0.312} 
 & 0.403 & 0.435 & 0.286 
 & \underline{0.306}  & \underline{0.338} & \underline{0.244}  \\

 DialoRPT~\citep{gao2020dialogrpt} 
 & -0.066$^*$ & -0.044$^*$ & -0.031$^*$ 
 & 0.267 & 0.244 & 0.166 
 & -0.077$^*$ & -0.069$^*$ & -0.049$^*$  \\
 
 GRADE~\citep{LishanHuang2020GRADEAG}
 & \underline{0.491} & 0.434 &0.300 
 & \underline{0.549} & \underline{0.568} & \underline{0.398} 
 & -0.031$^*$ & -0.005 & -0.030$^*$ \\

QuantiDCE(R)~\citep{quantdce}
 & 0.348 & 0.300 & 0.202
 & 0.498 & 0.507 & 0.351
 & 0.162 & 0.182 & 0.130 \\

QuantiDCE(P)~\citep{quantdce}
 & 0.408 & 0.387 & 0.234
 & 0.522 & 0.521 & 0.372
 &  0.238   & 0.257& 0.189   \\

 QuantiDCE(F)~\citep{quantdce}
 & 0.458 & 0.427 & 0.265
 & 0.589 & 0.577 & 0.436
 &  0.278 & 0.326 & 0.237   \\
\hline

\rowcolor{Gainsboro} \multicolumn{10}{l}{\textit{Reference-based lexicon-level methods} }  \\

ROUGE-1~\citep{lin-2004-rouge}       
&    \hllemon{0.217}      &  0.192   & 0.133 &  
\hllemon{0.221} & \hllemon{0.217}  & \hllemon{0.151} 
& 0.116 & 0.101 & 0.069   \\

ROUGE-2~\citep{lin-2004-rouge}              
&    0.210       &  0.145   &  0.148  
&  0.009$^*$   & 0.046  &  0.058 
& 0.065 & 0.040 &0.032  \\

ROUGE-L~\citep{lin-2004-rouge}
& 0.215 & 0.178 &0.129 
& 0.213 & 0.214 & 0.148 
& {0.118} & { 0.114} & {0.079}   \\

BLEU-1 ~\citep{papineni2002bleu}       
&  0.201       &  0.190  &0.131 
&  \hllemon{0.115}  &  0.118  & 0.076 
& 0.010 & 0.081 & 0.055     \\

BLEU-2  ~\citep{papineni2002bleu}             
&  0.201    &  0.200 & 0.158   
&  0.057  &  0.041$^*$   & 0.032 
& 0.060 & 0.039 & 0.031    \\

BLEU-3 ~\citep{papineni2002bleu}             
 &  0.201         &  0.189   & 0.153 
 &  0.049  &  0.036 & 0.030$^*$  
 & 0.017 & -0.001$^*$ & -0.001$^*$  \\
 
BLEU-4  ~\citep{papineni2002bleu}             
&  0.203         &  \hllemon{0.207}  & 0.169  
&  0.059  &  0.056 &   0.046 
& 0.017 & -0.005$^*$ & -0.004$^*$  \\

METEOR~\citep{banerjee2005meteor}
& 0.202 & 0.188 &0.129 
& 0.182 & 0.194 & 0.132 
& 0.099 & 0.051 & 0.035  \\ 

\hline
\rowcolor{Gainsboro} \multicolumn{10}{l}{\textit{Reference-based semantic-level methods} }  \\
 
 Bertscore$_\text{p}$~\citep{TianyiZhang2019BERTScoreET}            
 & 0.347     &  0.334    &  0.334 
 &  0.229 &0.146&  0.104 
 & -0.446 & -0.089 & -0.061$^*$  \\
 
 Bertscore$_\text{r}$~\citep{TianyiZhang2019BERTScoreET}              
 &  0.296     & 0.243    &  0.213  
 &  0.167 & 0.243 & 0.173 
 & 0.278  & 0.292 & 0.196   \\
 
 Bertscore$_\text{f1}$~\citep{TianyiZhang2019BERTScoreET}              
 &  0.229        &  0.308    &  0.213  
 &  0.211  &0.204  & 0.145 
 & 0.133 & 0.115 & 0.079  \\
 
 BARTScore ~\citep{MichaelLewis2019BARTDS}
 & 0.133 & 0.057 & 0.039 
 & 0.256 & 0.253 & 0.173 
 & 0.143 & 0.168 & 0.115  \\
 
 RUBER~\citep{ChongyangTao2017RUBERAU} 
& 0.332 & 0.351 & \underline{0.369}
& 0.252 & 0.256 & 0.183 
&0.122 & 0.123 & 0.089   \\

 BLEURT~\citep{sellam2020bleurt} 
 & 0.353 & 0.363 & 0.249 
 & 0.343& 0.337& 0.232 
 & 0.105 & 0.140 & 0.102  \\
 
BERT$_{\text{MLP}}$$^\dagger$  ~\citep{bert}  
 &    0.241       &  0.255  &0.173    
 &  0.186  & 0.225  & 0.153 
 &  0.274 & 0.330 & 0.202    \\
 
BERT$_\text{MLP}$$^\dagger$~\citep{bert}
&    0.304       &  0.301   &0.192   
&  \underline{0.501} & \underline{0.537} & \underline{0.373} 
&  \underline{0.331} & \underline{0.360} & \underline{0.251}   \\

Roberta$_\text{MLP}$$^\dagger$    ~\citep{roberta}
&  0.275 &  0.306 &0.300 
&  0.285  & 0.307 & 0.307 
& 0.317 & 0.334 & 0.223    \\
 
BART$_\text{MLP}$$^\dagger$ ~\citep{MichaelLewis2019BARTDS}   
&  \underline{0.431} &   \underline{0.440} & 0.312 
&   0.412 & 0.447 & 0.356  
&  0.310 & 0.335 & 0.242   \\

\hline
\rowcolor{Gainsboro} \multicolumn{10}{l}{\textit{Reference-assisted methods}} \\
\textbf{RADE} (Pre-trained model, PT) 
& 0.472 & 0.491 & 0.334
& 0.650 & 0.601 & 0.427 
& 0.386 & 0.390 & 0.285  \\ 

\textbf{RADE} (Task-specific model, TS) 
& \hllemon{\textbf{0.601}} & \hllemon{\textbf{0.569}} & \hllemon{\textbf{0.409}} 
& \hllemon{\textbf{0.863}} & \hllemon{\textbf{0.849}} & \hllemon{\textbf{0.685}} 
& \hllemon{\textbf{0.470}} & \hllemon{\textbf{0.465}} & \hllemon{\textbf{0.347}} 
 \\ 
\bottomrule
\end{tabular}
}
\caption{\textbf{Details} of our automatic evaluation experiment on three benchmarks.
We divide the methods in Reference-free, Reference-based and RADE, while the reference-based including the semantic-level and lexicon-level methods.
Note that $r$, $\rho $ and $\tau$ indicate the Pearson's $\rho$, Spearman's $r$ and Kendall'$\tau$. 
All values are statistically significant to p-value < 0.05, unless marked by$^*$. 
The FED$_\text{E}$ and FED$_\text{U}$ indicate two evaluation perspective of FED, i.e., engagement and understandability.
Methods with $^\dagger$ are implemented by ourselves.
We \underline{underline} the best results of each group of baselines methods and \textbf{bold} the best results of all methods.
}
\label{automatic_all_eval}
\end{table*}

\begin{table*}[!t]
\small
\centering
\begin{tabular}{@{}c lccccc @{}}
\toprule
\textbf{Topic}                                       
& \multicolumn{6}{p{14cm}}{Library of Alexandria}
\\
\midrule
\multirow{3}{*}{\textbf{Context}}                       
& \multicolumn{6}{p{14cm}}{\underline{User1}: The library of Alexandria had a unique way of gathering books by making all ships visiting give their books for copy. }  \\
& \multicolumn{6}{p{14cm}}{\underline{User2}: They must have had a impressive collection!} \\
& \multicolumn{6}{p{14cm}}{\underline{User1}: How unfair, they would copy their books give them back the copy and keep the originals.} \\ 
\midrule
\textbf{Reference}                                       
& \multicolumn{6}{p{14cm}}{ I guess that is true. Do you think we will ever have a centralized digital library of all our works and knowledge?}
\\ \cmidrule(l){2-7} 
\textbf{Response}                                       
& \multicolumn{6}{p{14cm}}{That is exactly why they didn't stick around and stay put. I wish I could work somewhere where I could share the workload.}
\\ \cmidrule(l){2-7} 
\multirow{2}{*}{\textbf{\textbf{Evaluation}}}   
& \multicolumn{1}{c}{Human (1-5)}          & QuantiDCE (1-3)          & GRADE (0-1)         & METEOR (0-1)         & BERTScore (0-1)         & RADE (1-5)           \\ \cmidrule(l){2-7} 
\multicolumn{1}{l}{}                            
& \multicolumn{1}{c}{3.67}                   & 3.00             & 0.26          & 0.17           &       0.48            & 3.54            \\
\bottomrule
\end{tabular}
\caption{An example of the DSTC dataset evaluated by five automatic and human evaluation methods. This example shows that our methods correlate more strongly with human judgment than the other methods.}
\label{case:1}
\end{table*}

\begin{table*}[!t]
\small
\centering
\begin{tabular}{@{}clccccc@{}}
\toprule
\textbf{Emotion}                                       
& \multicolumn{6}{p{14cm}}{Confident}  \\
\midrule
\multirow{3}{*}{\textbf{Context}}                       
& \multicolumn{6}{p{14cm}}{\underline{User1}:I love when I need to show someone how to do something and everything goes off without a hitch.}     \\
& \multicolumn{6}{p{14cm}}{\underline{User2}:Oh yeah, that is always nice. Sometimes they just do not get it, though.}  \\
& \multicolumn{6}{p{14cm}}{\underline{User1}:Fortunately, it is usually just not having the best answer for what they ask. I have learned and prepared a lot more this time.} \\ 
\midrule
\textbf{Reference}                                       
& \multicolumn{6}{p{14cm}}{Wow, you sound pretty confident! You must really know what you are talking about.}                                            
\\ \cmidrule(l){2-7} 
\textbf{Response}                                       
& \multicolumn{6}{p{14cm}}{I agree, I am sure you will do great things.}                                                                                   \\ \cmidrule(l){2-7} 
\multirow{2}{*}{\textbf{Evaluation}}   
& \multicolumn{1}{c}{Human (1-5)}          & QuantiDCE (1-3)          & GRADE (0-1)         & METEOR (0-1)         & BERTScore (0-1)         & RADE (1-5)               \\ \cmidrule(l){2-7} 
\multicolumn{1}{l}{}                            
& \multicolumn{1}{c}{3}                   & 3.00              & 0.45          & 0.04        &      0.57         & 3.64                   \\
\bottomrule
\end{tabular}
\caption{An example of the EmpatheticDialogue dataset evaluated by five automatic and human evaluation methods.
}
\label{case:2}
\end{table*}

\begin{table*}[!t]
\small
\centering
\begin{tabular}{@{}clccccc@{}}
\toprule
\textbf{Emotion}                                       
& \multicolumn{6}{p{14cm}}{Excited}  \\
\midrule
\multirow{3}{*}{\textbf{Context}}          
& \multicolumn{6}{p{14cm}}{\underline{User1}: I am looking forward to my best friend s surprise birthday party next week!}                               \\
& \multicolumn{6}{p{14cm}}{\underline{User2}: That sounds like so much fun! I love parties!} \\
& \multicolumn{6}{p{14cm}}{\underline{User1}: I am really happy about it. She is a great friend and she is turning 40, so it is a big one!} \\ 
\midrule
\textbf{Reference}                                       
& \multicolumn{6}{p{14cm}}{Hey, I just had that one! What do you have planned?}  \\ 
\cmidrule(l){2-7} 
\textbf{Response}                                       
& \multicolumn{6}{p{14cm}}{That is great. Do you have any other day?}                                                                                   
\\ \cmidrule(l){2-7} 
\multirow{2}{*}{\textbf{Evaluation}}   
& \multicolumn{1}{c}{Human (1-5)}          & QuantiDCE (1-3)          & GRADE (0-1)         & METEOR (0-1)         & BERTScore (0-1)         & RADE (1-5)           \\ \cmidrule(l){2-7} 
\multicolumn{1}{l}{}                            
& \multicolumn{1}{c}{2.98}                   &   2.7             & 0.29          & 0.34        &       0.61          & 2.64            \\
\bottomrule
\end{tabular}
\caption{An example of the EmpatheticDialogue dataset evaluated by five automatic and human evaluation methods. 
}
\label{case:3}
\end{table*}

\begin{table*}[!t]
\small
\begin{tabular}{@{}clccccc@{}}
\toprule
\multirow{3}{*}{\textbf{Personality}}                       
& \multicolumn{6}{p{14cm}}{(1): I like to donate time and money to charity.} \\
& \multicolumn{6}{p{14cm}}{(2): I work as a computer programmer.} \\
& \multicolumn{6}{p{14cm}}{(3): i make over 200 , 000 per year.  } \\
\midrule
\multirow{5}{*}{\textbf{Context}}                       
& \multicolumn{6}{p{14cm}}{\underline{User1}: Hi, how are you ? tell me something about yourself!} \\
& \multicolumn{6}{p{14cm}}{\underline{User2}: Well, I love going fishing with my four attractive daughters every day.} \\
& \multicolumn{6}{p{14cm}}{\underline{User1}: Sounds fun! I enjoying volunteering and donating to charities.   } \\ 
& \multicolumn{6}{p{14cm}}{\underline{User2}: Cool maybe you'd like to run a charity at my new race track. I race cars!} \\
& \multicolumn{6}{p{14cm}}{\underline{User1}: Sounds exciting! I am a computer programmer, which pays over 200k a year.} \\ 
\midrule
\textbf{Reference}                                       
& \multicolumn{6}{p{14cm}}{Would you like to marry one of my four attractive daughters? I will sell one.}  \\ 
\cmidrule(l){2-7} 

\textbf{Response}                                       
& \multicolumn{6}{p{14cm}}{wow! That's a lot of money. Do you have any hobbies?}                                                                                   \\ \cmidrule(l){2-7} 
\multirow{2}{*}{\textbf{Evaluation}}   
& \multicolumn{1}{c}{Human (1-5)}          & QuantiDCE (1-3)          & GRADE (0-1)         & METEOR (0-1)         & BERTScore (0-1)         & RADE (1-5)          \\ \cmidrule(l){2-7} 
\multicolumn{1}{l}{}                            
& \multicolumn{1}{c}{3.25}                   &  2.87              & 0.02         & 0.06           &      0.53           & 2.91           \\
\bottomrule
\end{tabular}
\caption{An example of the PersonaChat dataset evaluated by five automatic and human evaluation methods, showing that our methods have a stronger correlation with human judgment than the other methods.}
\centering
\label{case:4}
\end{table*}

\begin{table*}[!ht]
\small
\begin{tabular}{@{}clccccc@{}}
\toprule
\multirow{3}{*}{\textbf{Personality}}                       
& \multicolumn{6}{p{14cm}}{(1): I like cupcakes and reading.} \\
& \multicolumn{6}{p{14cm}}{(2): I work at a grocery store.} \\
& \multicolumn{6}{p{14cm}}{(3): My favorite color is blue.  } \\ 
\midrule
\multirow{5}{*}{\textbf{Context}}                       
& \multicolumn{6}{p{14cm}}{\underline{User1}: Hi! what are you up to ? I am doing what I like best, eating cupcakes!} \\
& \multicolumn{6}{p{14cm}}{\underline{User2}: Hi I am winding down from a day at work.} \\
& \multicolumn{6}{p{14cm}}{\underline{User1}: So am I. The local grocery store is where I work. What about you?  } \\ 
& \multicolumn{6}{p{14cm}}{\underline{User2}: I also work in the retail industry at the local department store!} \\
& \multicolumn{6}{p{14cm}}{\underline{User1}: Other than eating cupcakes, reading is also what I like to do to wind down. } \\ 
\midrule
\textbf{Reference}                                       
& \multicolumn{6}{p{14cm}}{I like to read also and play with my dog. Do you have a pet?}  \\ 
\cmidrule(l){2-7} 
\textbf{Response}                                       
& \multicolumn{6}{p{14cm}}{What do you do for fun? My girlfriend and I go to the lake a lot.} \\ \cmidrule(l){2-7} 
\multirow{2}{*}{\textbf{\textbf{Evaluation}}}   
& \multicolumn{1}{c}{Human (1-5)}          & QuantiDCE (1-3)          & GRADE (0-1)         & METEOR (0-1)         & BERTScore (0-1)         & RADE (1-5)           \\ \cmidrule(l){2-7} 
\multicolumn{1}{l}{}                            
& \multicolumn{1}{c}{2.75}                   &   3.00         & 0.01       & 0.22          &      0.58         & 2.79              \\
\bottomrule
\end{tabular}
\caption{An example of the PersonaChat dataset evaluated by five automatic and human evaluation methods. This example shows that our methods have a stronger correlation with human judgment than the other methods.}
\centering
\label{case:5}
\end{table*}

\end{document}